\documentclass[conference,compsoc]{IEEEtran}
\usepackage{stmaryrd}%
\usepackage{amsfonts}
\usepackage{url}

\usepackage{nicefrac}

\usepackage{graphicx,amsmath} 

\usepackage{tikz}
\usetikzlibrary{arrows,shapes,shapes.symbols,shapes.geometric,positioning,automata,calc,trees,positioning,chains,decorations.pathmorphing,decorations.pathreplacing,matrix}

\IEEEoverridecommandlockouts    

\begin{document}





\title{Learning and Inferring Relations in Cortical Networks}%

\author{\IEEEauthorblockN{
Peter U. Diehl\IEEEauthorrefmark{1},
Matthew Cook\IEEEauthorrefmark{1}
}
\IEEEauthorblockA{\IEEEauthorrefmark{1}Institute of Neuroinformatics\\
ETH Zurich and University Zurich, Switzerland\\ 
Email: peter.u.diehl@gmail.com}
\thanks{This work was supported by SNF Grant 200021-143337.}%
}



\maketitle

\begin{abstract}
A pressing scientific challenge is to understand how brains work.
Of particular interest is the neocortex,
the part of the brain that is especially large in humans,
capable of handling a wide variety of tasks including
visual, auditory, language, motor, and abstract processing.
These functionalities are processed in different self-organized regions of the neocortical sheet,
and yet the anatomical structure carrying out the processing
is relatively uniform across the sheet.
We are at a loss to explain, simulate, or understand such
a multi-functional homogeneous sheet-like computational structure --
we do not have computational models which work in this way.
Here we present an important step towards developing such models:
we show how uniform modules of excitatory and inhibitory neurons
can be connected bidirectionally in a network
that, when exposed to input in the form of population codes,
learns the input encodings
as well as the relationships between the inputs.
STDP learning rules lead the modules
to self-organize into a relational network,
which is able to infer missing inputs,
restore noisy signals,
decide between conflicting inputs,
and combine cues to improve estimates.
These networks show that it is possible for
a homogeneous network of spiking units
to self-organize so as to
provide meaningful processing of its inputs.
If such networks can be scaled up,
they could provide an initial computational model relevant to
the large scale anatomy of the neocortex.
\end{abstract} 

\section{Introduction}

Despite immense effort to understand and model the mammalian neocortex, it remains a challenge to engineer neural networks whose architecture and functionality both approach that of the neocortex.

In terms of functional properties, probably the best models available include deep learning \cite{lecun2015} and probabilistic frameworks \cite{ghahramani2015}, which have shown dramatic improvements in recent years and even match human performance in a range of tasks including visual object recognition and face verification.
However, although there seem to be similarities in the learned representations used by deep neural networks and those in visual areas of the macaque monkey brain \cite{yamins2014}, it is clear that there are profound differences between deep learning and cortical processing.

Regarding architecture,
models attempting to capture the detailed anatomy of the neocortex have as yet been limited in their ability to do general learning \cite{fregnac2014, heinzle2007}.
However, models that simply capture the rough statistics of neocortex,
for example having groups (``areas'') of spiking units consisting of
80\% excitatory neurons and 20\% inhibitory neurons with randomized connectivity,
have succeeded in capturing abilities such as reliable memory recall of neuronal assemblies and the formation and maintenance \cite{litwin2014, zenke2015}.
The neocortex consists of many such areas, bidirectionally connected,
but such a macro-architecture has not previously been shown to be able to learn and make inferences.

Here we present a spiking neural network model whose
design was constrained simultaneously by
the top-down requirement that it should be able to
solve relational inference tasks
and by
the bottom-up requirement that it should be based on
biologically relevant architectures and plasticity mechanisms.
In this way, the model serves to connect
the biological implementation level
with
the computational goal of learning and inference,
helping to bridge the gap between these levels of Marr's hierarchy~\cite{marr1976}).

Two forms of
spike-timing-dependent plasticity (STDP)
are used, one for excitatory synapses and one for inhibitory synapses.
As in neocortex,
excitatory neurons
project both locally and inter-areally,
and outnumber by a factor of four the inhibitory neurons,
which only project locally.
The more detailed laminar structure of neocortex is not modeled.
A spiking neuron model is used whose parameters
have been set to match electrophysiological data.

The network yields useful functionality such as
soft winner-take-all \cite{oster2009, maass2000, riesenhuber1999}
and signal restoration \cite{Jug12}.
These networks learn the statistics of their input,
enabling them to perform tasks such as
coordinate transformation and inference propagation,
and they fit well with biological experimental data on cue integration \cite{ernst2006, knill2004, gu2008} 
and invariance of the response tuning-width to different input strengths \cite{sclar1982, chao1984, skottun1987}.

The network is built from identical modules, which are
capable of
processing incoming information
to infer the values of any missing or noisy information.
Such modules are then connected bidirectionally,
as observed in the cortex,
yielding effective relational inference.

The framework of relational networks provides a powerful tool
for thinking about neural networks on a more abstract level,
using populations of neurons as basic building blocks
which can learn to process complex input structures
not just in a feedforward manner but in an omnidirectional way:
After a relation is learned
it can then be used to
``solve'' for any single variable of the relation in terms of the others.
This enables a relation to perform inference
on any variable of the relation
rather than just a selected output variable,
so the network can perform a chain of inference,
starting with whatever is known (or observed),
and proceeding to any conclusions that can be reached
from the available information.

\begin{figure}[t]
\includegraphics[width=3.3in]{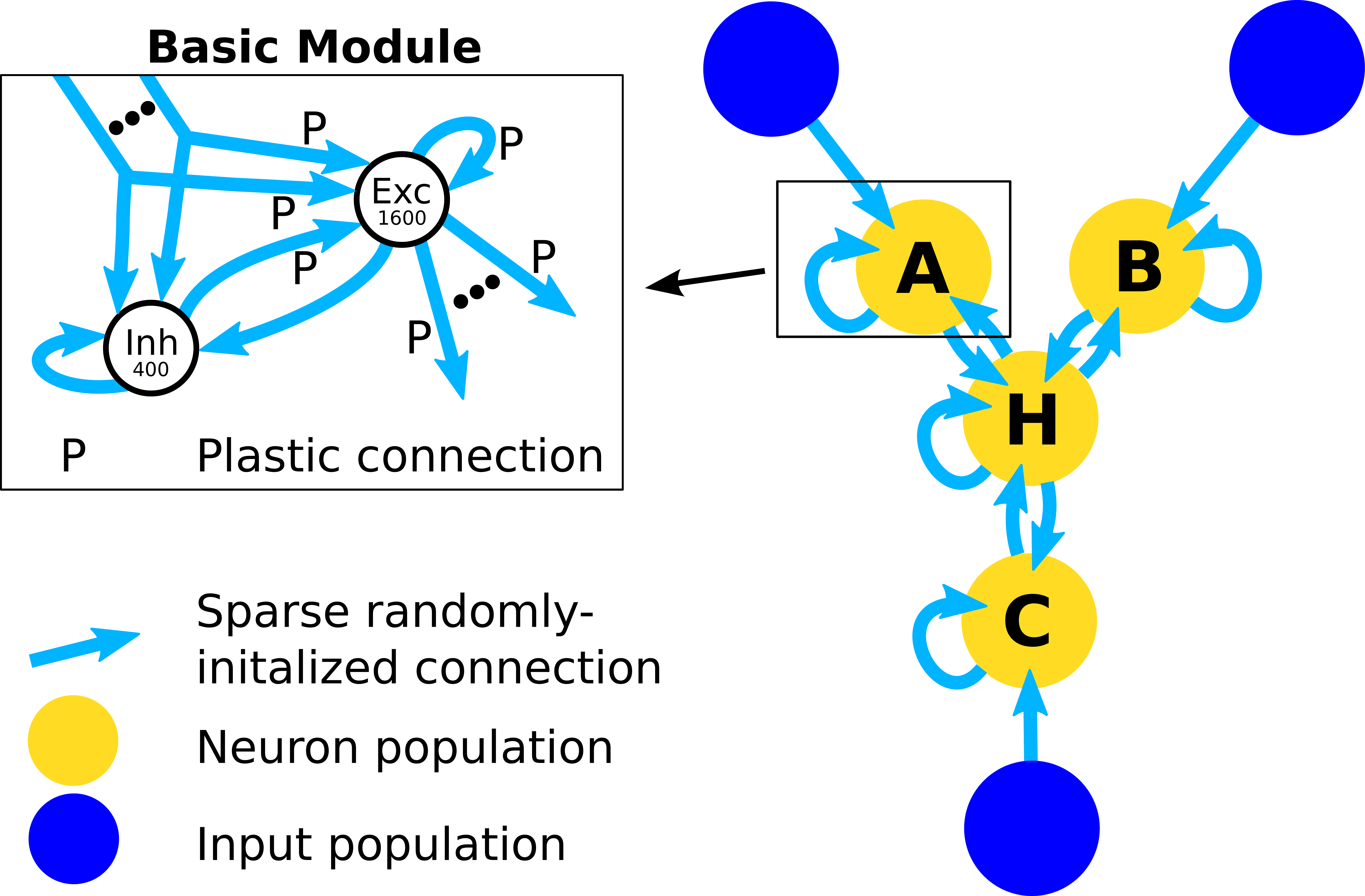}
\caption{\textbf{Architecture of the network.} 
The network is composed of generic modules:
input populations (shown in blue) and neuron populations (shown in yellow).
Input populations consist of 1600 excitatory axons,
bringing Poisson-distributed spike-trains
whose rate is set according to the data item being observed.
The internal structure of the neuron populations
is shown in the inset.
They are composed of 1600 excitatory and 400 inhibitory leaky integrate-and-fire neurons. 
Within a neuron population, all possible types of connections are formed and all connections received by excitatory neurons are plastic (using STDP).
Neuron populations are coupled using bidirectional long-range connections originating from excitatory neurons and targeting excitatory and inhibitory neurons.
\textit{All} connections are sparse (10\% of possible connections are made) and the weights are randomly initialized.
}
\label{fig:architecture_3way}
\end{figure}

\begin{figure}[h!]
\includegraphics[width=3.3in]{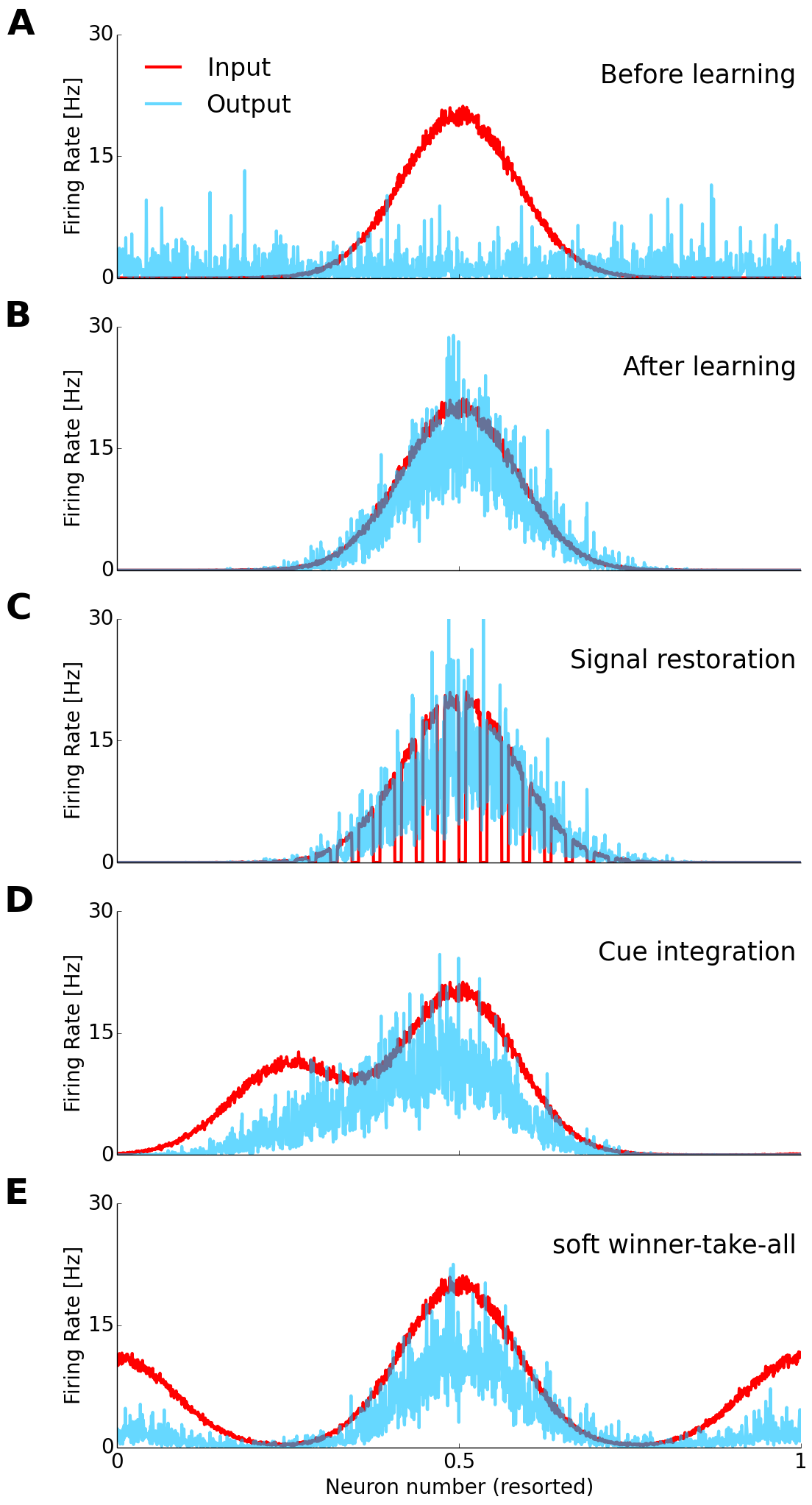}
\caption{\textbf{Response patterns of a single population.} 
Inputs are shown in red and population activity (after neuron sorting) is shown in blue.  
A) Response before learning.
B) Response after learning has converged (30'000 examples). All following graphs also show properties of the network after learning.
C) Signal restoration. The activity of 32\% of the input neurons is set to 0 (see red downward strokes). The recurrent connections restore the desired shape and prevent a loss of activity in the parts where input is missing.
D) Cue integration. Additionally to the main stimulus (with peak at 0.5), another similar input is present (with peak at 0.25). 
The resulting activity is close to the main stimulus but is also biased towards the additional input.
E) Soft winner-take-all. As in D), just with stimuli suggesting completely different values (peaks are at 0 and at 0.5). 
The resulting activity is dominated by the position of the stronger stimulus and the weaker stimulus is suppressed.
}
\label{fig:responsePatterns}
\end{figure}

\section{Results}
In order to successfully learn a large-scale inference network, we need single modules, our basic building blocks, to fulfill properties such as a large spectrum of possible input strengths, contrast invariance of the tuning curves, soft winner-take-all, cue-integration, and signal restoration.
In the first part of this section we will show how a single module behaves in regard to those properties and in the second part we build larger networks using those basic blocks.
The second part describes results for simulating networks with 4 modules that learn the relations $A+B-C=0$,  $A = 0.5 + B = -C$ and $A = 2 B = C^2$.
It is important to note that all of the networks use the same initial structure to learn the relations.
The architecture is shown in figure \ref{fig:architecture_3way}.
Finally, we show that the 3-way network is capable (after learning) of inferring missing variables for the different types of relations, again without requiring any changes in the initial structure of the network.

\begin{figure*}
	\includegraphics[width=0.5\textwidth]{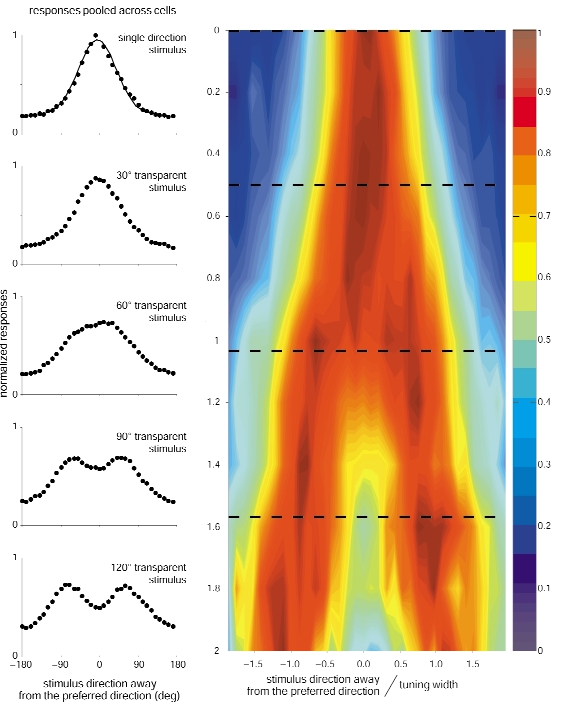}
	\includegraphics[width=0.5\textwidth]{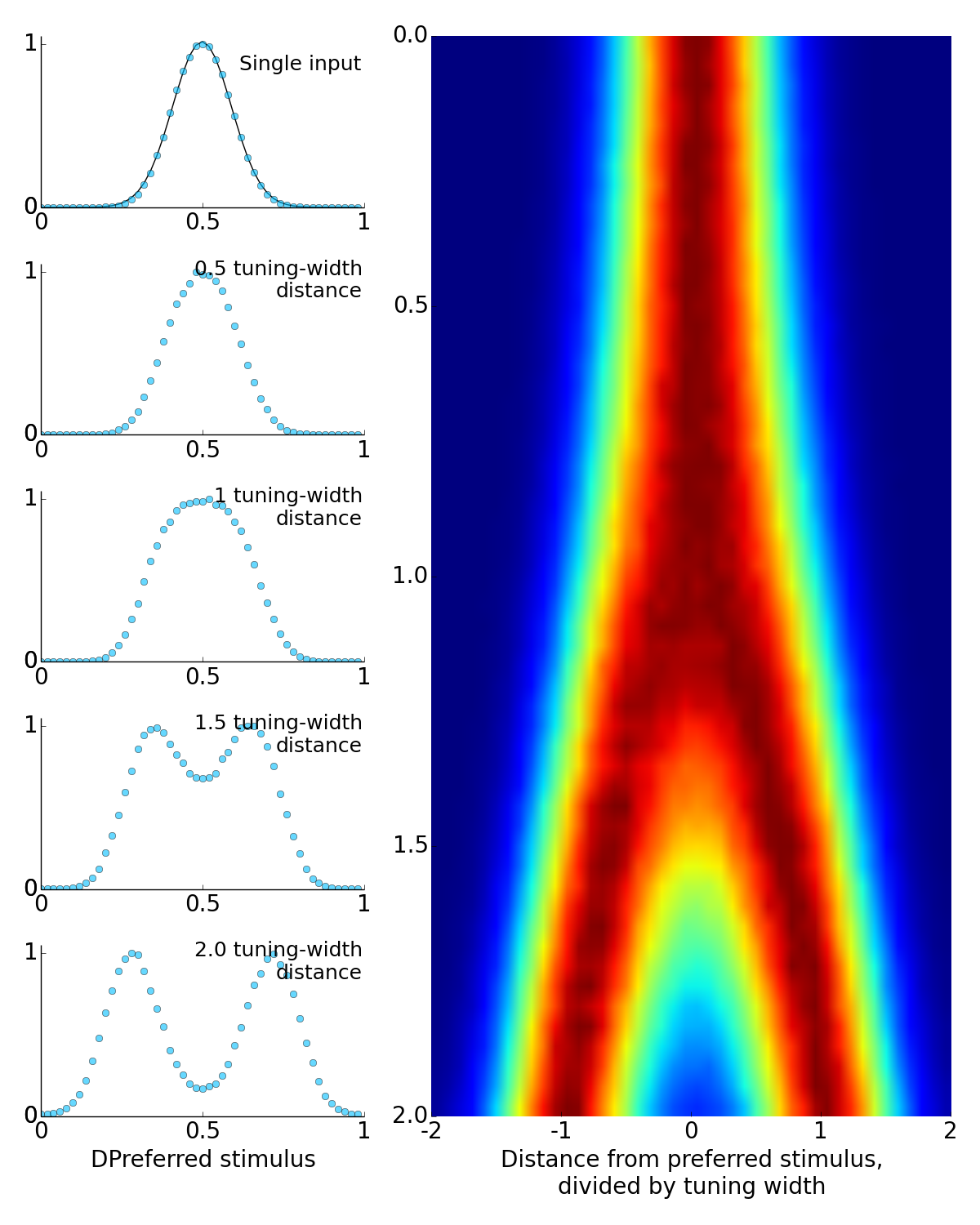}
    \caption{\textbf{Perceiving multiple directions of motion.}
        The left part of the figure depicts recordings of direction-selective neurons in area MT of the macaque monkey (reproduced from \protect\cite{treue2000}).
        Specifically, it shows the population activity profiles for dot motion of various angles.
        The right part of the figure shows results of population responses from simulations of a basic module.
        Both parts follow the same structure with five response patterns for different angles on the left and and the gradual change of the response for different angles on the right.
        For the observed data as well as for the simulation, the responses begin to split when the distance from the preferred stimulus is greater than one tuning width.}
\label{fig:multipleAngles}
\end{figure*}

\subsection{Single Population Properties After Learning}
In this part we show the properties of a single neuron population as depicted in figure \ref{fig:architecture_3way}.
The standard input are Poisson-distributed spike-trains with a Gaussian-shaped firing rate profile as it has been observed in experiments \cite{hubel1962}, see the red curve in figure \ref{fig:responsePatterns} A.
Before learning the response of the network is (as to be expected) very broad (blue curve in \ref{fig:responsePatterns} A).
However, after learning of the network has converged, the response of the population clearly reflects that it learned to represent the input, see figure \ref{fig:responsePatterns} B.

In the following we show how the population responds to different types of inputs (which it was not trained on) to test for its tolerance to noise, errors, and ambiguity.
Note that all properties shown are a result of the neural mechanisms and the network architecture we used, i.e. none of the input patterns that are used for testing population properties was used during training. 


\subsubsection{Signal Restoration}
Natural stimuli and systems are inherently noisy, which means a plausible system should be able to deal with missing, incomplete or distorted inputs. 
In figure \ref{fig:responsePatterns} C is shown how a trained network deals with noisy input.
Even though 32\% of the inputs are set to 0, the resulting output activity of the population matches its desired shape and does not shown a loss of activity for the parts where inputs are missing.
This is possible because of the recurrent connections which after learning match the input to the closest known patterns and create a combination of the "own belief" (the memories stored in the recurrent connections) and the current information about the environment (represented by the incoming input).

\subsubsection{Cue Integration}
If the population receives two similar but different stimuli, the resulting response of the population is an average of the two stimuli that is weighted by the strength of the inputs, see figure \ref{fig:responsePatterns} D.

This behaviour is similar to what intuitively is expected (i.e., when two (non-conflicting) inputs are shown that hint towards different solutions) it is reasonable to assume a weighted average of them \cite{ernst2006, knill2004, gu2008}.

\subsubsection{Ambiguity Resolution}
In order to have some form of decision making on a single population level, the network should exhibit competition between conflicting inputs. 
Specifically, if there is a conflict between two inputs (i.e., inputs which suggest very different values) the stronger one should dominate and reduce the activity of the lower one in a non-linear fashion.
This is also called winner-take-all (WTA), and if it is not enforced that exactly one solution wins and the other is of but if a compromise is allowed it is called soft WTA. 
The response of a single (learned) population is shown in figure \ref{fig:responsePatterns} E.
Although the sum of inputs increases, the network does not show a stronger response which is desirable since conflicting inputs should not increase the belief in either one of them. 
Note that it can be seen that the response to the smaller stimulus is indeed suppressed when comparing the observed response under competition to the input-output relation without competition (see section  \ref{sec:input-output} for more details).

\subsubsection{Multiple Inputs}

In this part we are using the same single population setup as in the last section but with different inputs.
Instead of encoding one value using one Gaussian-shaped input, we are using two Gaussian-shaped input profiles to represent two (potentially) different values. 

Specifically, in the event of multiple inputs for the same variable, there are two possibilities to process the inputs, they can be combined as shown in the subsection 'cue integration' or they can be process as two separate inputs. When there is equal evidence for both of them (and not one of the stimuli is dominating as in the cue integration and the soft-winner-take-all case) the stimuli will be averaged if they are close enough and will be perceived as two distinct stimuli.

In \cite{treue2000} the authors used two random dot stimuli with a varying angle between their movement directions.
Those two random dot stimuli are then displayed on top of each other and if the angle between the dot movements is close enough, the stimuli are perceived as one.
At a certain angle this combination of the movements will vanish and the two patterns will be perceived as two separate surfaces moving.
This separation starts happening when the angle between the two movement directions is bigger than the orientation tuning-width of the orientation selective neurons \cite{treue2000}. 

One interesting property of our network is to determine the conditions under which the transition from combination to separation occurs. 
When applying two equally strong stimuli with different values to the same population we can compare the experimentally predicted separation angle of the two different patterns to the separation angle of our model, see figure \ref{fig:multipleAngles}.
Our model matches the experimental findings that if the difference between the two inputs is bigger than one standard deviation of the single stimulus response, the response pattern changes from a single peak to two peaks.

\begin{figure}[h!]
\includegraphics[width=3.3in]{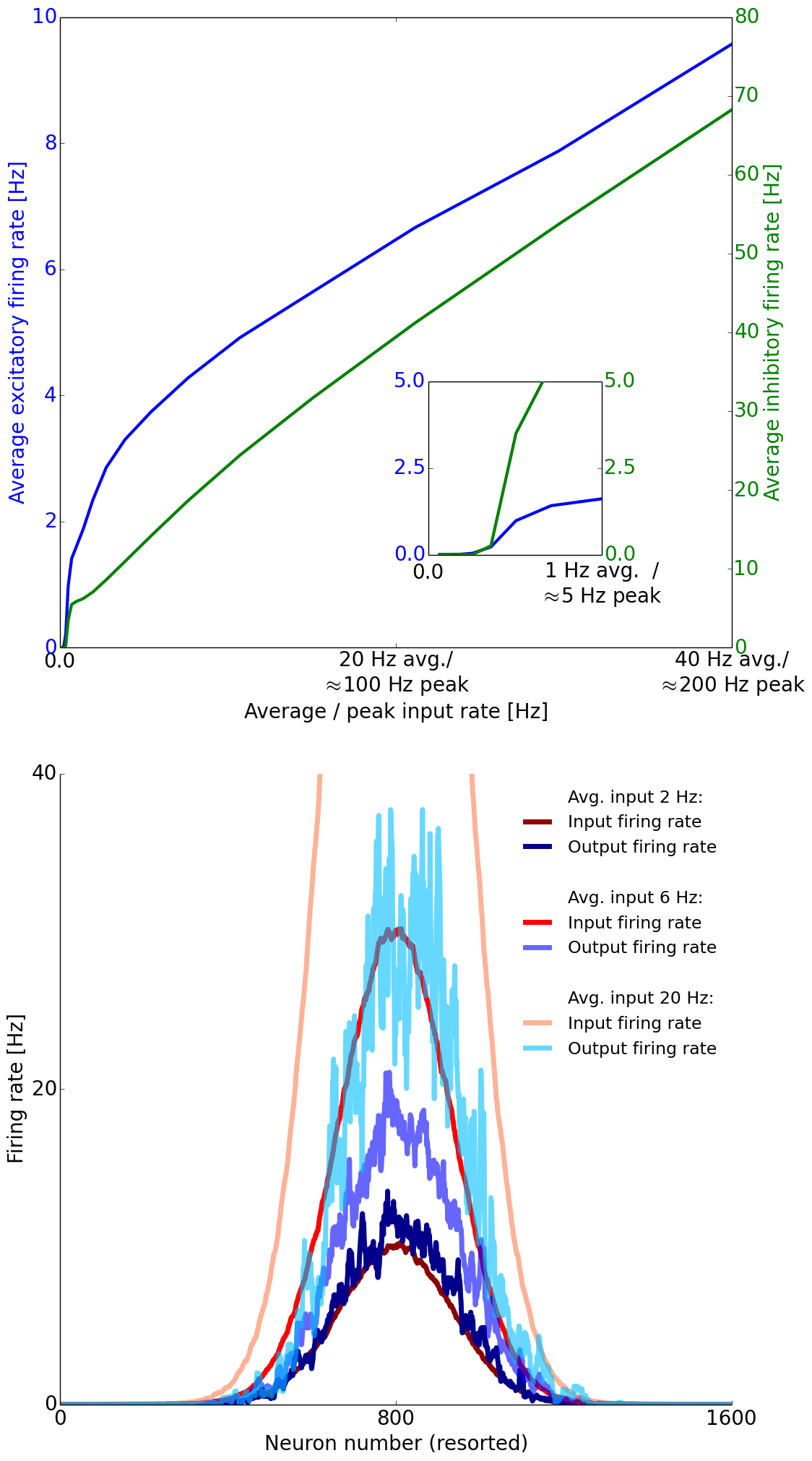}
\caption{\textbf{Input-output relationship and response-width invariance.} 
Upper plot: 
The x-axis shows the average/peak input firing rate in Hz and the y-axis shows the average (output) firing-rate of the receiving excitatory neurons (shown in on the left side in blue) and the average firing-rate of the receiving inhibitory neurons (right side in green).
Note that while the activity of the excitatory neurons is approximately Gaussian distributed (as shown in the lower plot), the activity of the inhibitory neurons is almost uniformly distributed.
Lower plot: 
The plot shows the firing rates of input neurons and excitatory neurons in the receiving population for three different input strengths (2 Hz, 6 Hz, and 20 Hz of average input firing rate).
On the x-axis are (sorted) neuron numbers and on the y-axis the firing rate The upper part of the 20 Hz input is not shown for clarity; the peak is at $\approx$ 100 Hz.    
All firing rates are averaged across neighboring neurons with a sliding window of size 10 to make the differences between the three setups more visible. 
This plot shows that the width of the output activity does not increase with increasing input strength.
A similar response is observed in visual cortex when presenting stimuli with different contrasts: the tuning width of the cells does not change depending on the contrast \protect\cite{sclar1982, chao1984, skottun1987}.
}
\label{fig:input_output}
\end{figure}

\subsection{Dependence of Output to Input Firing-rate and Response-width Invariance}\label{sec:input-output}

The brain must not only deal with different input intensities but also with entirely missing inputs, which requires the system to be very adaptable to overall input strength. 
Therefore, for many practically relevant tasks, a simulated neural network should be able to deal with the same variance in input strength.

The simulation setup to test this property is comparable to the previous two, i.e. we are using a single population with Gaussian-shaped inputs.
However, this time for each example we use a single Gaussian-shaped input with varying amplitudes of the Gaussian.
In figure \ref{fig:input_output} we show the summed activity of all neurons in a trained population as a function of the input strength.
Specifically, inputs as low as 2 Hz elicit a response of about 1 Hz in the postsynaptic excitatory neurons, while increasing the input by a factor of 50 leads to an increase of less than a factor of ten.
Therefore network can deal with widely varying input strengths without suffering from an explosion of activity.
This is an especially exciting property since many networks already struggle with doubled input strengths and since it will be necessary for building larger scale networks which receive multiple inputs and need to be able to deal with incomplete or even missing inputs.

It's been known for a long time that neurons in visual cortex do not change the width of their Gaussian-shaped responses with changing contrast of the input \cite{sclar1982, chao1984, skottun1987}. 
This phenomenon is known as "contrast invariance of the tuning width" and can also be observed in our model, see lower part of figure \ref{fig:input_output}.
Even for a ten-fold increase in input firing rate, the width of the response stays nearly the same.

\begin{figure*}
\includegraphics[width=7.0in]{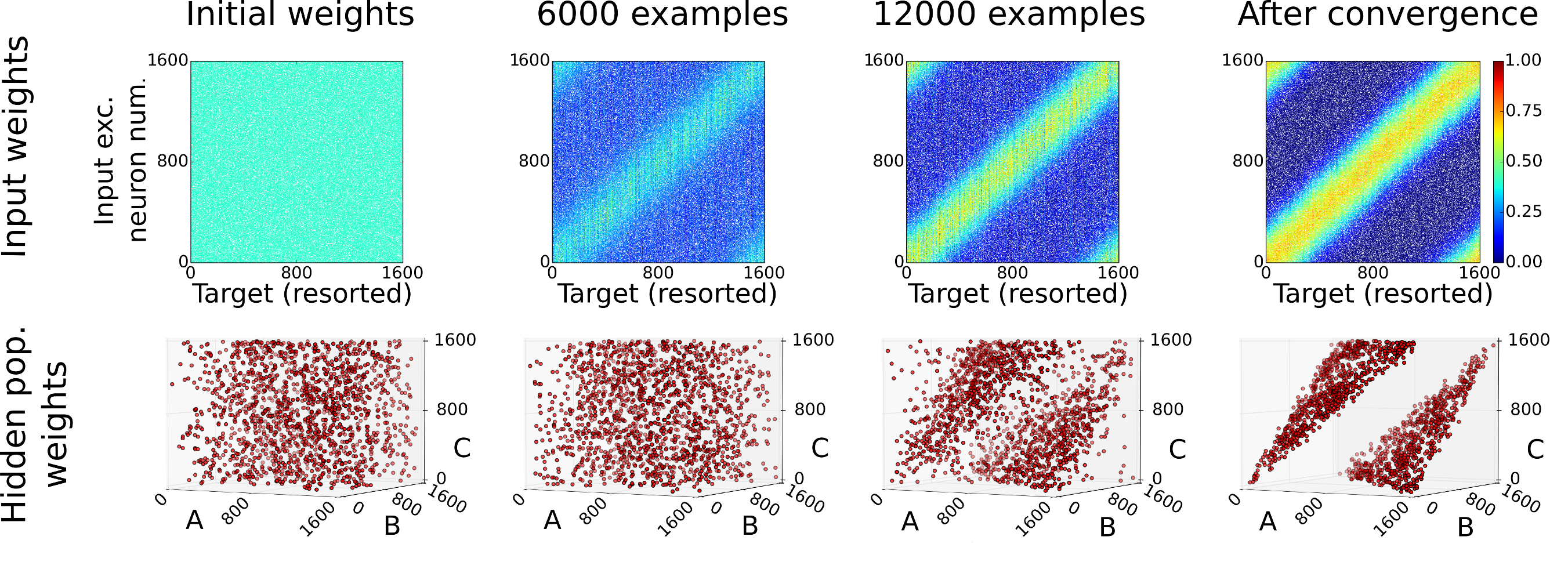}
\caption{\textbf{Weight development during learning of the relation $A+B-C=0 \ mod \ N$. }
Top row: Connection matrices from input population to neuron population A. The neurons are resorted according to their circular average connection strength from the input population.
Bottom row: 3D-plot of the weights for each neuron in $H$. The three axes represent the circular average connection strength to each of $A$, $B$, and $C$ (after their neurons are resorted according to input tuning). The four shown states are before learning, after 6,000 examples, after 12,000 examples, and after convergence (30,000 examples).
}
\label{fig:weightDevelopment}
\end{figure*}

\begin{figure*}
\includegraphics[width=7.0in]{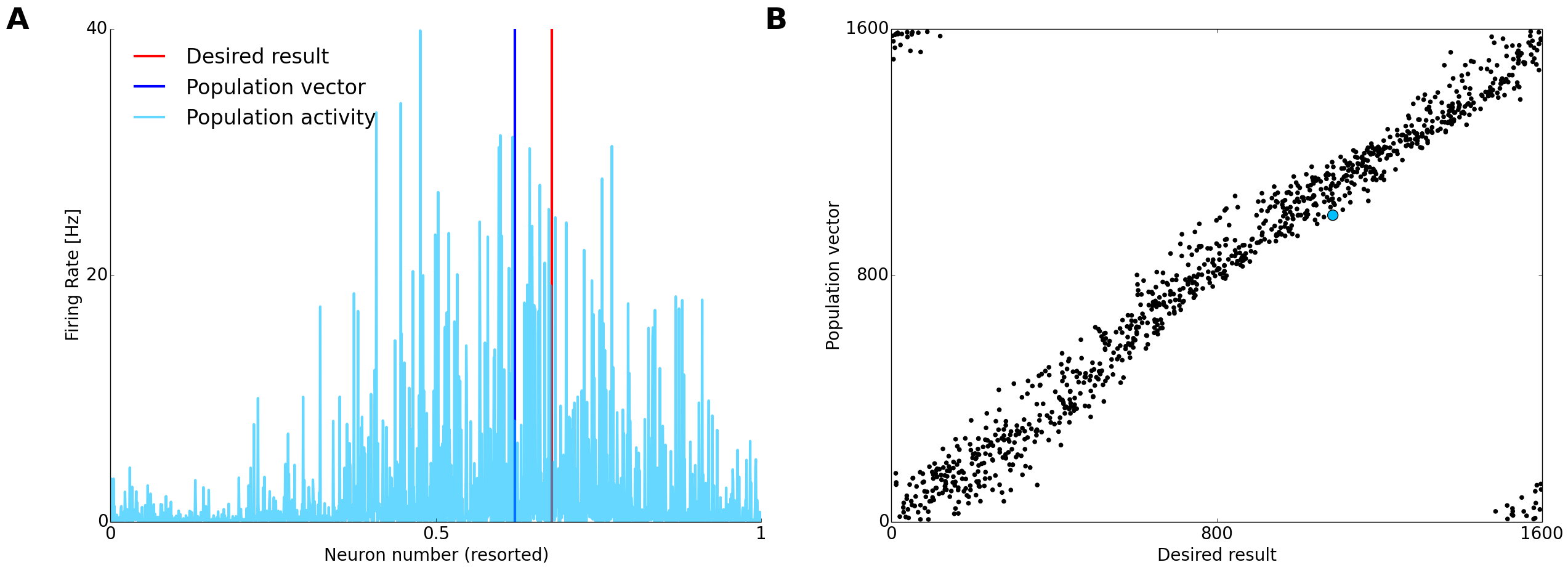}
\caption{\textbf{Inferred activity after learning.} 
The population was trained with examples satisfying the relation $A+B-C=0 \ mod \ N$.
After learning, input was provided to the populations $A$ and $B$ but not to $C$.
Note that this would work the same way when, instead of not providing input to $C$, no input would be provided to either $A$ or $B$. 
$B$) The plot shows the response of population $C$ to a single example (i.e., a specific input provided only to $A$ and $B$).
The x-axis denotes the (sorted) neuron number and the y-axis denotes the firing rate of each neuron during 60 seconds of stimulus presentation.
The desired result ($A+B \ mod \ N$) is depicted by a red vertical bar and the inferred value (i.e. the actual population vector of $C$) is depicted by a blue vertical bar.
B) Shown are the inferred values of 1000 randomly drawn inputs for $A$ and $B$.
For each one of the examples, one black dot is shown with the x-axis denoting the desired result ($A + B$) and the y-axis denoting the population vector.
The ideal solution would be the identity.
The dot of the population vector and desired result of the example shown in A) is colored cyan.
Note that this was one of the worse examples in terms of discrepancy between population vector and desired result.
Each of the examples was presented for 1 second.
}
\label{fig:singleInference}
\end{figure*}

\begin{figure*}
	\includegraphics[width=0.33\textwidth]{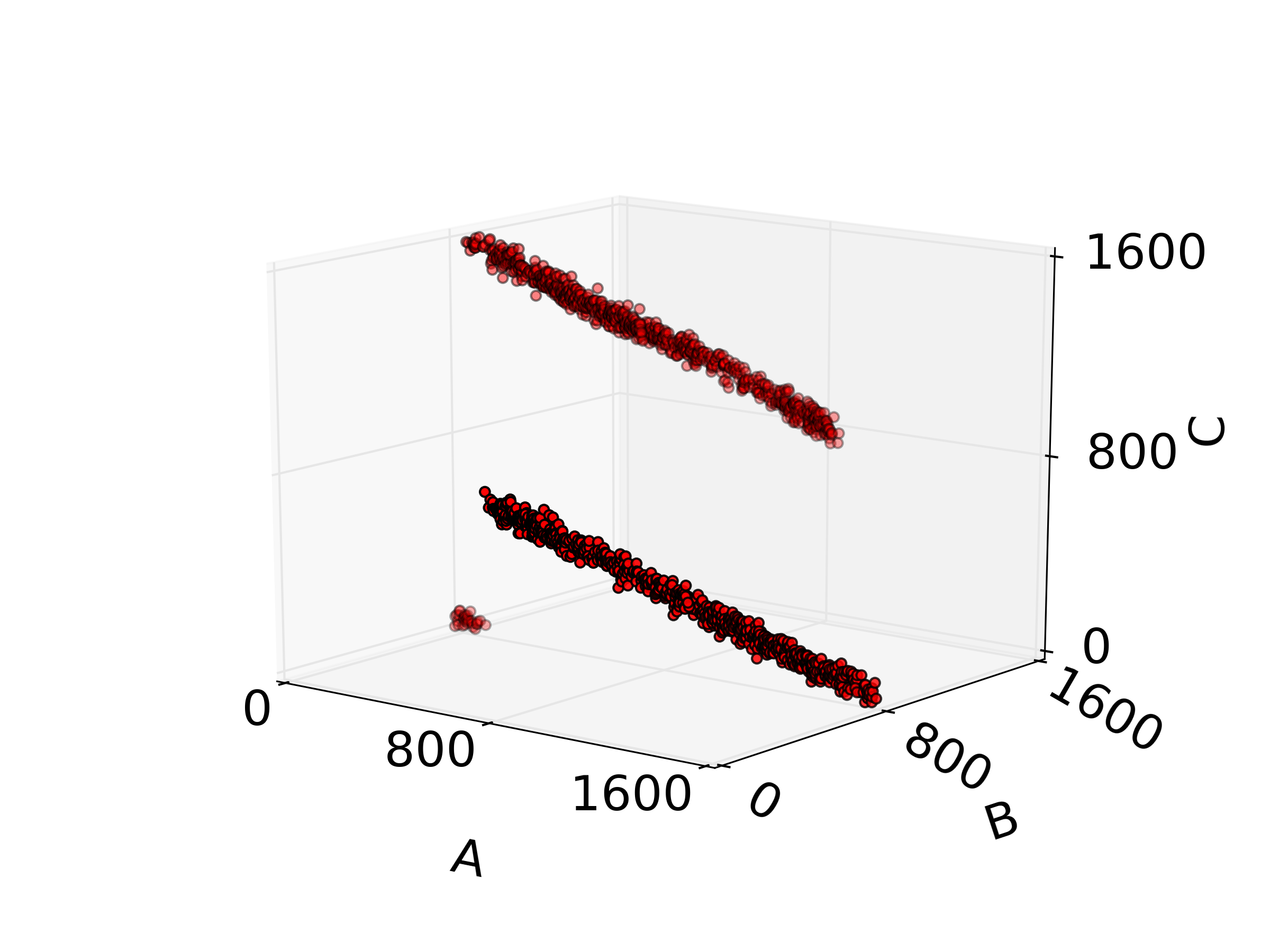}
	\includegraphics[width=0.33\textwidth]{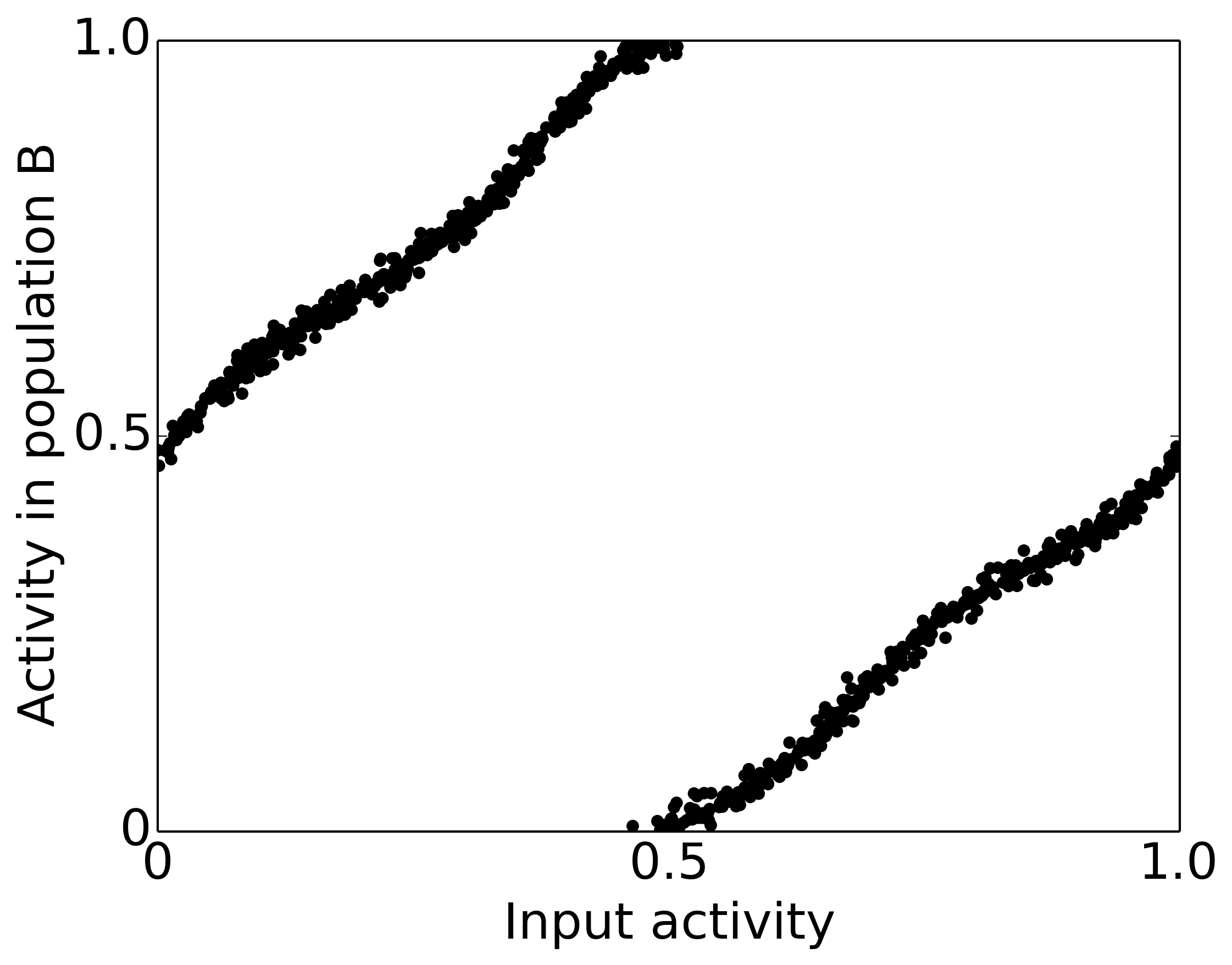}
	\includegraphics[width=0.33\textwidth]{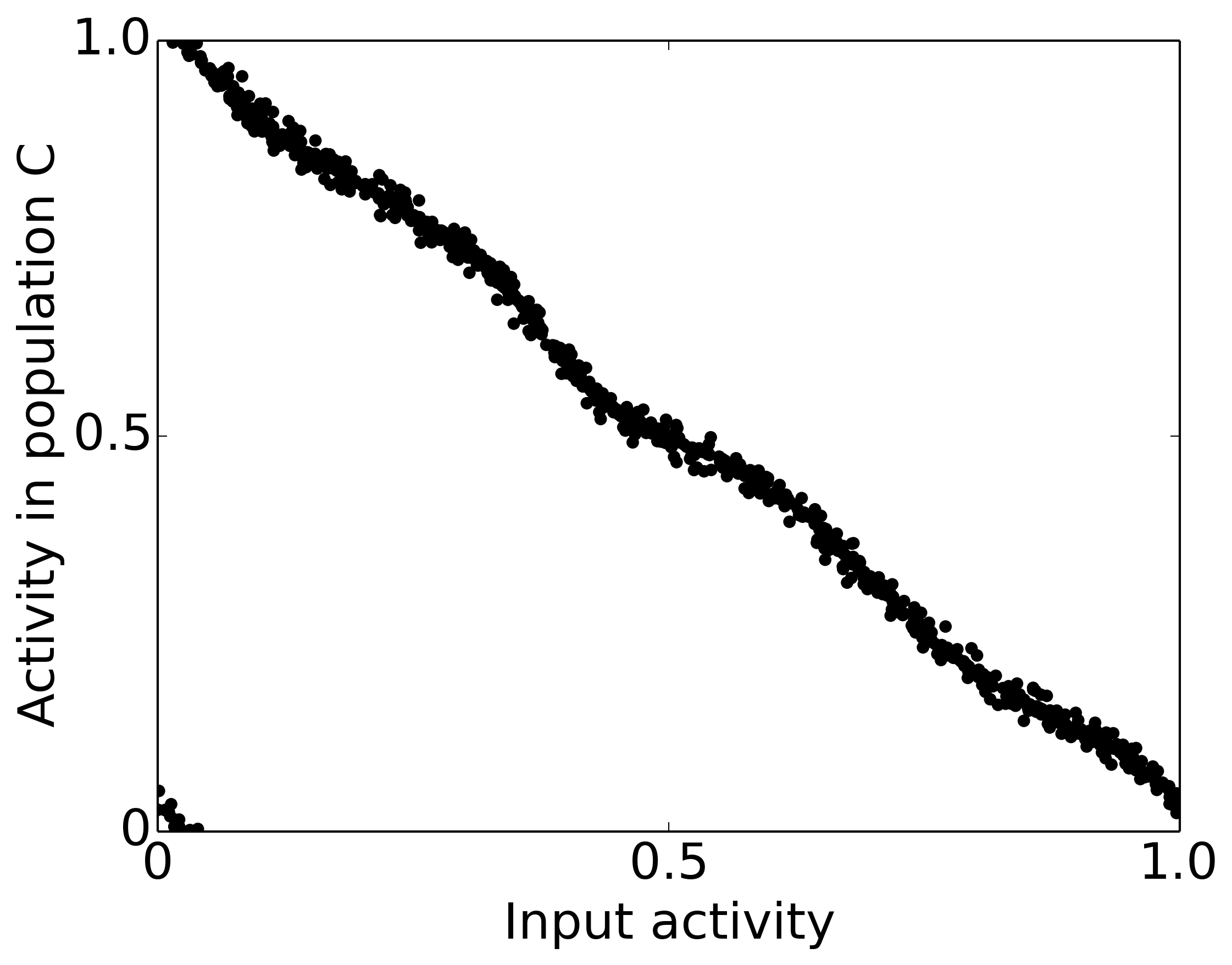}
    \caption{\textbf{Learned weights and inference of the relation $A = 0.5 + B = -C$.}
    The left plot shows the weights of the hidden population $H$, see bottom row of figure \protect\ref{fig:weightDevelopment} for more details. 
    The middle and the left plots show (similar to the right plot in figure \protect\ref{fig:singleInference}) the responses of the network to 1000 input examples.
    An input is provided to population $A$ but not $B$ or $C$, instead the activity in both $B$ and $C$ is inferred from $A$.
    Therefore, the two plots on the middle and right show the inferred activity from $A$ in $B$ and $C$, respectively.
    }
\label{fig:negative}
\end{figure*}

\begin{figure*}
	\includegraphics[width=0.33\textwidth]{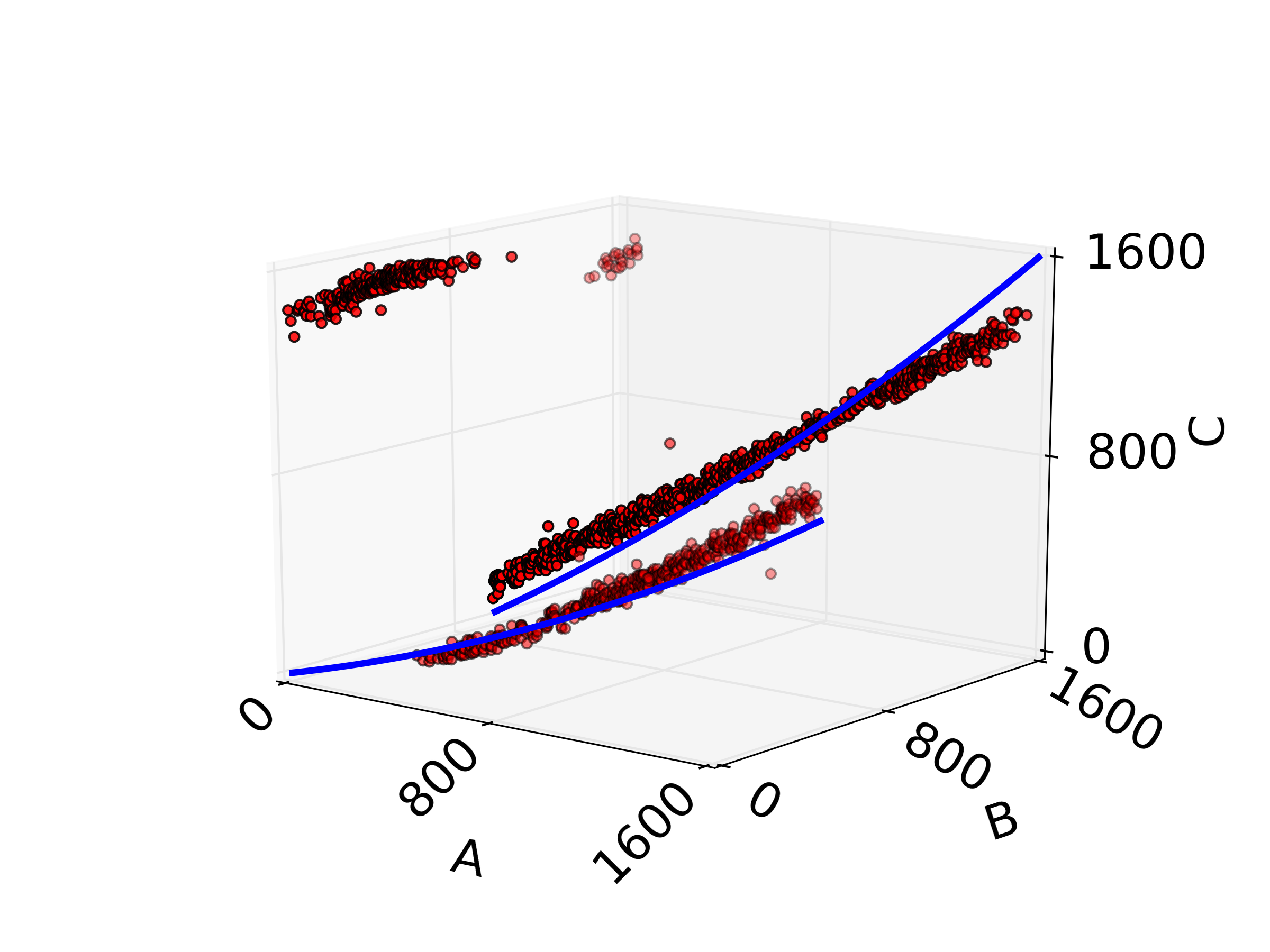}
	\includegraphics[width=0.33\textwidth]{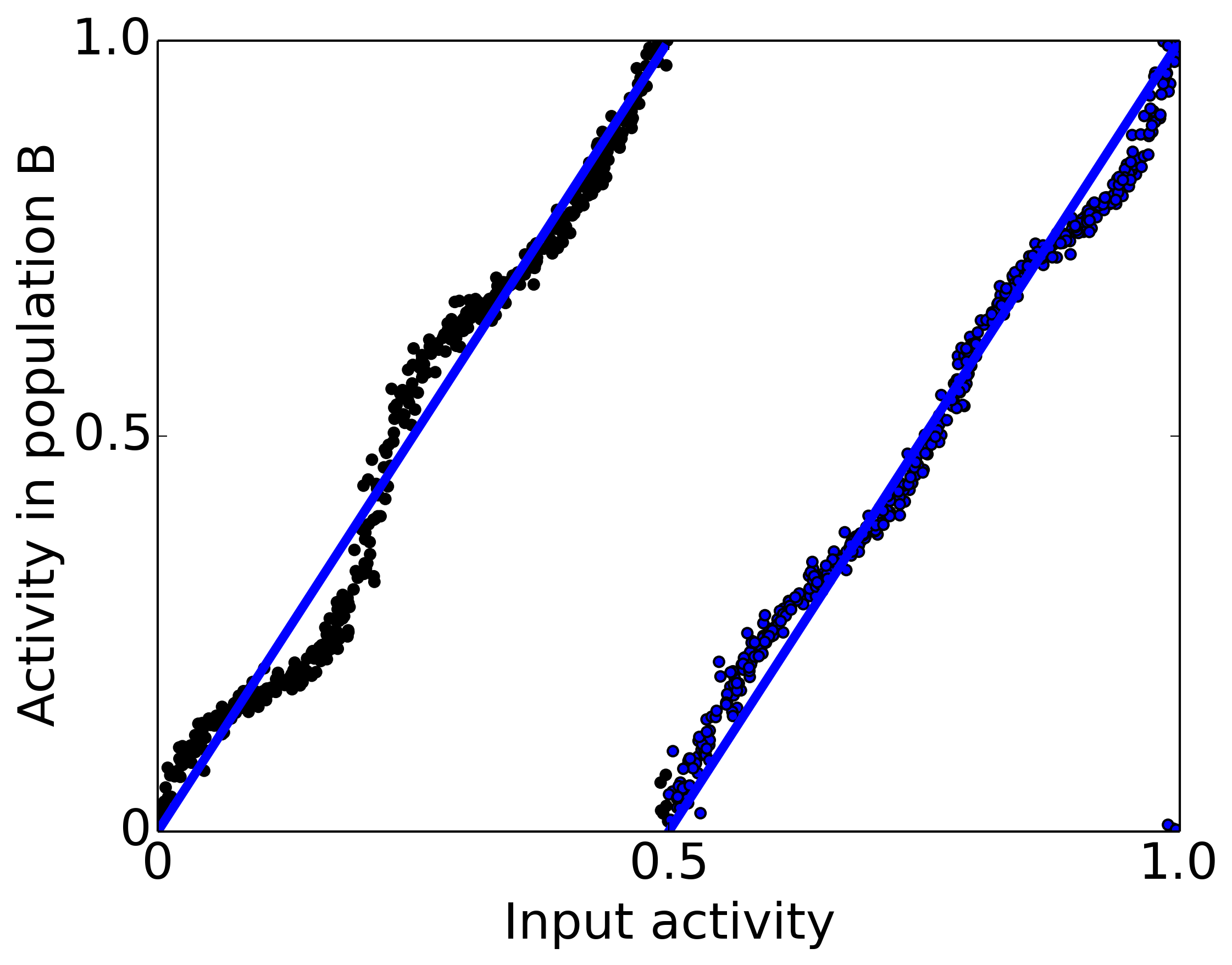}
	\includegraphics[width=0.33\textwidth]{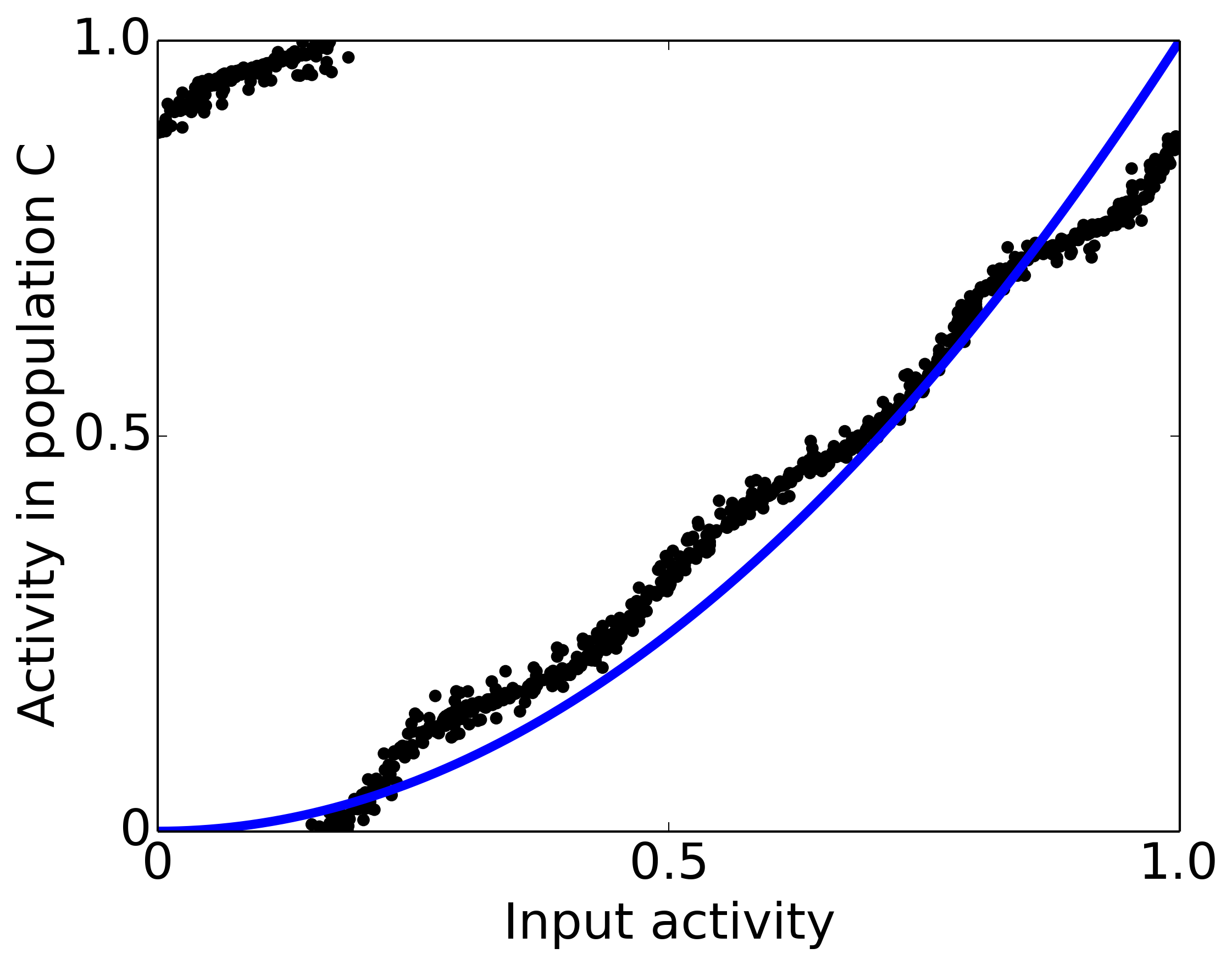}
    \caption{\textbf{Learned weights and inference of the relation $A = 2 B = C^2$.} 
    Visualization is identical to the one figure \protect\ref{fig:negative}.
    Due to the wrap around at if $B>1$ or $B<0$, each value in $B$ can be caused by two different inputs. 
    To show where the wrap-around should occur, blue dots are used in the middle figure.}
\label{fig:doubleSquare}
\end{figure*}

\subsection{Learning and Inferring Relations}

In order to be able to learn and infer relations, we need to build networks that can handle multiple inputs.
The simplest case for a network with multiple inputs is a two-way relation, i.e. a network that associates two variables with each other.
By combining multiple such two-way relations, it is possible to create chains of relations and therefore chains for reasoning.
If we add one more population to the network, thereby creating a three-way network, this can be used as a building block for arbitrarily complex systems.
For relating $n$ variables to each other, it is only necessary to connect $n-2$ three-way relations.
Therefore we will go on to describe how to build such a three-way network.

The three-way network consists of three input populations (mainly for simulating realistic input to the network) and four neuron populations for processing, see figure \ref{fig:architecture_3way}.
Three of the four neuron populations are processing the inputs and the remaining ("hidden") population receives inputs from the other three, enabling the hidden population to learn the relation between the three (pre-processed) inputs.
By connecting the neuron populations in a bidirectional way, it is possible to learn relations between the different entities instead of just the function (one-way).
Therefore, after learning it is possible infer any missing inputs instead of just the output from the input as it is typically the case in neural networks. 

If one or more inputs are missing, the network infers the missing ones. 
Naturally, the effect of this inference is only useful if the connections of the network contain necessary knowledge for the inference. 
Since we initialize all connections randomly, the only way to acquire the capability to correctly infer missing variables is via learning.
The learning between populations occurs via in the same way as for the connections within a population, i.e. via spike-timing dependent plasticity (STDP).

\subsubsection{Adding Variables}
The first example relation we will use to demonstrate the learning capabilities of the network is $A + B - C = 0$.
It is not only an interesting example which is applicable to real-life situations \cite{Deneve01} but it can also be visualized well, see figure \ref{fig:weightDevelopment}.
The top row shows the weight matrices from an input population to a neuron population (note that since all three inputs are encoded the same way the matrices will look very similar but not identical due to different input examples).
The bottom row of figure \ref{fig:weightDevelopment} shows the weights of the hidden population $H$ to the other three neuron populations.
Specifically, each red dot corresponds to one neuron in $H$ and its position in the 3D space is determined by its strongest weights to $A$, $B$, and $C$.

Learning appears to happen in two stages: first the three outer populations learn the encoding of the input and after that the hidden population learns the relation between the three inputs.
As it can be seen in figure \ref{fig:weightDevelopment}, learning of the input relation corresponds to a rearrangement of the neurons in $H$.
Since the relation we picked is a 2D relation, all neurons are drawn towards this 2D sheet during learning.
However, in principle neurons could be arranged in any pattern in this 3D space, which again corresponds to the ability to learn any relation between the three variables including nonlinear ones (two more examples of other relations are described below).

\subsubsection{Adding Constants and Negating}
Given the fact that the network is able to learn to add different variables, it is no surprise that it can learn to add constants as well.
In fact, it can do so without an additional input for the constant.
Moreover, the network can also learn to negate a given input.
Figure \ref{fig:negative} shows a network that learned both of those relations for different variables (i.e., it learned $A = 0.5 + B = -C$).
What might come as a surprise from the traditional perspective of functional systems is that (after learning) our relational system is able to infer \textit{two} missing variables given only one.
The two plots on the right of Figure \ref{fig:negative} show the inferred activity of $B$ and $C$ given only $A$.

\subsubsection{Squaring and Doubling}
All relations presented so far were linear ones. 
However, in principle the network does not differentiate between linear and nonlinear ones.
The main challenge for nonlinear ones is that when the relation implies a big change for a small change in the inputs, more neurons are needed to represent the relation well for those values.

As examples of nonlinear relations we chose multiplication and raising the input to a power.
Specifically, the relation between the inputs is $A = 2 B = C^2$, see the left plot of Figure \ref{fig:doubleSquare} for the resulting weights of the central population.
Similarly to the relations of the last network (adding constants and negating), it is possible to infer two missing values given only one input.
The middle and right plot of Figure \ref{fig:doubleSquare} show the activity of populations $B$ and $C$ during inference.
Due to the wrap-around the relation $A = 2B$ implies that the value of $B$ when $A=0.25$ is equal to the value of $B$ when $A=0.75$.

\begin{figure*}
\includegraphics[width=7.0in]{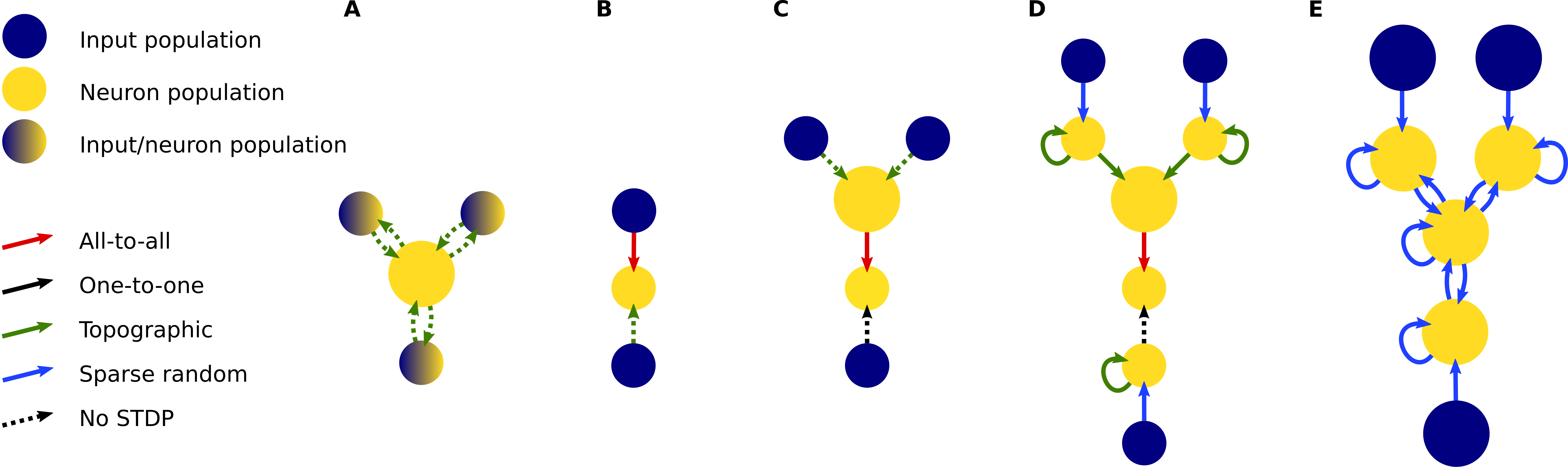}
\caption{ \textbf {Network architectures} from:
A) Deneve, Latham, Pouget \cite{Deneve01}, 
B) Davison and Fregnac 2006 \cite{Davison06}, 
C) Wu et al. 2008 \cite{Wu08},
D) Srinivasa and Cho 2012 \cite{Srinivasa12} and
E) our implementation.
Circles denote neuron populations and arrows their connections.
For the implementations B) and C) only one output layer is shown.
Note that the previous networks B) to D) are feedforward (besides recurrent connections within a population in D)) which enables them to learn functions but not relations.
Also, all of the previous networks use hardwired connections or learn with supervision (i.e., use static connections which are hardwired correctly to teach another connection what the correct activity of a neuron is). 
}
\label{fig:architectures}
\end{figure*}

\section{Discussion}

\subsection{Previous Work}

Architectures which are similar to the proposed one have been studied in different contexts such as cue integration, decision making and coordinate transformation \cite{Davison06, Deneve01, Srinivasa12, Wu08}.
However, previous networks were either hardwired or were missing feedback connections (see Figure~\ref{fig:architectures}). 
Additionally all previous architectures used some non-plastic connections to help learn the remaining ones and even the learned connections often already started with a suitable topographic connectivity. 
By employing plastic connections
for such topographic connectivity
researchers were able to check whether
their network could
maintain such a state without degenerating. 
Nonetheless, our main goal was to show that
it is possible for the network to learn
{\em all} connections which are received
by excitatory neurons
and thereby train a truly self-organizing network.

\subsection{Relational Paradigm}

Our network is designed around what could be called a
relational paradigm.
The type of processing envisioned is not the standard
feed-forward paradigm of an output determined by the inputs,
but is rather understood as
a collection of semi-independent areas,
each embodying some local knowledge,
working together to
converge into a globally consistent state.

For example, in the network used in this paper,
the population $A$ contains knowledge
about how the input activity patterns
are related to the relayed value
sent between $A$ and $H$.
Meanwhile, $H$ contains knowledge about
what are valid combinations of relayed values
from $A$, $B$, and~$C$.
In general,
the bidirectional connection between two populations
serves to get those populations to agree on
the value of a latent variable,
and each population contains the knowledge of how
the various latent variables that it sees
are related,
much like the Forney normal form of a factor graph.

This has many advantages over
traditional function approximation architectures.
For example,
it is very general due to its
``tabula rasa'' \cite{kalisman2005} approch,
which allows it to learn arbitrary population codes
(including the standard one-dimensional ones used here),
as no topology is ``hard-wired'' into the network.
Also, the network will infer any missing values,
not just the ones the network creator planned for in advance.
For coordinate transformations \cite{pouget1997} for example,
a network computing forward kinematics can simply be used in the other direction to compute inverse kinematics.
Other advantages include being able to
incorporate feedback to improve noisy output,
and the ease of adding new variables to a network
just by adding more modules,
since all variables are represented in the same way,
through learned responses in the modules.


A significant difference
between
our approach and previously presented architectures 
is that
each population sends its state
to all neighboring populations (except input populations)
and receives information from all neighboring populations,
in addition to sparse all-to-all recurrent connectivity
within each population.
In other words, every possible projection exists,
except between populations that are not neighbors
(in the sense of Figure~\ref{fig:architectures}).

\subsection{Challenges}
The fundamental difference in processing between feed-forward networks and highly recurrent networks does offer many advantages for processing and inference 
but this paradigm also presents certain challenges.
Recurrent connections within and (even more so) recurrent connections between populations can lead to undesired feedback loops.

When inspecting figures \ref{fig:negative} and \ref{fig:doubleSquare}, it becomes apparent that the inferred activity has some lumps while weights of the respective hidden populations appear to be beautifully straight.
Also. input weights are just as good as the ones shown in the upper row of figure \ref{fig:singleInference} (data not shown).
The reason for this difference between appearance of weights and of activity is that the recurrent connections amplify slight preference towards some solutions.

While the error resulting from those lumps does not pose a serious problem for the discussed three-way network, it might present a challenge when scaling up the architecture and using populations that are significantly farther away from direct input populations.
An easily implemented solution is to reduce the learning rate and avoid bumps by presenting more examples.
However, for practical applications it might be desirable to explore mechanisms that could prevent the network from creating too strong attractors towards certain solutions.

\section{Methods}
The simulations were done in BRIAN \cite{Goodman08}. The corresponding python scripts (including the used parameters) are available under {https://github.com/peter-u-diehl/threeway}.

We will start by describing a single neuron and then move on to describe how to construct a neuron population, i.e. one module.
Using this module as a basis, we will describe how to connect different neuron populations with each other to enable them to learn relations between them.
A list of the parameters can be found in tables \ref{tab:neuron_params}, \ref{tab:conn_params}, and \ref{tab:plast_params}.

\subsection{Neuron and Synapse Model}
The neuron model we employ is a (non-adaptive) leaky-integrate and fire (LIF) model described by the following equations
\begin{equation}
\tau \dfrac{dV}{dt} = (v^X_{rest} - V) + g^E (v^E_{g} - V) +  g^I (v^I_{g} - V)
\end{equation}
where $X$ is the neuron type and $V$ is the membrane voltage, see table \ref{tab:neuron_params} for the corresponding parameters.

The synapses are conductance based exponential synapses with an instantaneous conductance increase of $w_{ij}$, the moment a presynaptic spike from neuron $i$ arrives at neuron $j$. The temporal dynamics of the conductance $g_e$ are modelled by
\begin{equation}
\tau^X_{g} \dfrac{dg}{dt} = -g^X
\end{equation}
where $X$ denotes the synapse type (excitatory or inhibitory) and $\tau^X_{g}$ the time constant of the synapse.

\begin{table}
	\caption{\textbf{Neuron Parameters}}
    
	\centering
	\begin{tabular}{l l l}
		\textbf{Symbol} & \textbf{Description} & \textbf{Value} \\ \hline
		$v^E_{rest}$ & E resting potential & -65 mV\\
		$v^I_{rest}$ & I resting potential & -60 mV\\
		$v^E_{reset}$ & E reset potential & -65 mV\\
		$v^I_{reset}$ & I reset potential & -45 mV\\
		$v^E_{thresh}$ & E threshold potential & -52 mV\\
		$v^I_{thresh}$ & I threshold potential & -40 mV\\
		$\tau^E_{mem}$ & E membrane time constant & 20 ms\\
		$\tau^I_{mem}$ & I membrane time constant & 10 ms\\
		$\tau^E_{refrac}$ & E refractory period & 5 ms\\
		$\tau^I_{refrac}$ & I refractory period & 2 ms\\
	\end{tabular}\label{tab:neuron_params}
\end{table}

\subsection{Neuron Population}
The neuron and synapse model described in the previous subsection are used to construct populations of neurons.
Such neurons are grouped and connected to form neuron population, see table \ref{tab:conn_params} for the corresponding parameters.
Each population consists of 1600 excitatory neurons and 400 inhibitory neurons.
The ratio was chosen to match neuroanatomical findings \cite{gabbott1986}.

All combinations of types of neurons are connected by recurrent sparse random connections with a connection probability $p$ of 10\%.
The different types of recurrent connections serve different purposes:
\begin{enumerate}
\item excitatory to excitatory connections help to learn previously shown patterns and reinforce patterns with similar statistics,
\item excitatory to inhibitory connections foster the competition among the neurons and ensure that the excitatory neurons will respond to different stimuli,
\item inhibitory to excitatory connections balance the activity of the excitatory neurons (see subsection 'Homoeostasis')
\item and inhibitory to inhibitory connections help to maintain an asynchronous irregular firing pattern \cite{Brunel00}.
\end{enumerate}

The high number of neurons per module is necessary to average out discretization effects due to too few inputs.
To illustrate this, even with 2000 neurons per population and connections received from 3-4 modules, every neuron receives only 600-800 connections, only few of which will be active during a particular stimulus since they have an average firing rate of 3 Hz (controlled by a homoeostasis mechanism explained below).
This is an order of magnitude less inputs than a real cortical neuron receives \cite{binzegger2007}.

\begin{table}
	\caption{\textbf{Neuron Population Parameters}}
    
	\centering
	\begin{tabular}{l l l}
		\textbf{Symbol} & \textbf{Description} & \textbf{Value} \\ \hline
		$N^E$ & Number of E neurons  & 1600 \\
		$N^I$ & Number of I neurons  & 400 \\
		$p$ & Connection probability  & 0.1 \\
		$t_{exmpl}$ & Duration of single example presentation & 250 ms \\
		$g^E$ & Conductance of E synapses & n.a. \\
		$g^I$ & Conductance of I synapses  & n.a. \\
		$\tau^E_{g}$ & Time constant E synapses & 5 ms \\
		$\tau^I_{g}$ & Time constant I synapses & 10 ms \\
		$v^E_{g}$ & Reversal potential E synapses & 0 mV \\
		$v^I_{g}$ & Reversal potential I synapses & -85 mV \\
	\end{tabular}\label{tab:conn_params}
\end{table}

\subsection{Connecting Populations of Neurons}\label{sec:pop_conns}
Two neuron populations can be connected by creating 'long range' connections from the excitatory neurons of each of the two populations, to all the neurons of the other population.
The connections between the network populations are bidirectional on a population level, i.e. if one population connects to another it will also receive connections from the target population.
Note however that this does not hold true on the single neuron level, i.e. if one neuron connects to another it (likely) does not receive a connection from the same neuron.
This way of connecting the populations is also observed in biology \cite{felleman1991}, i.e. if one area connects to another than it also receives connections from there.
This 'coupling' of populations enforces that the connected populations reach a consistent state.

\subsection{Learning}
Learning between excitatory neurons is modelled by spike-timing dependent plasticity (STDP), see table \ref{tab:plast_params} for the corresponding parameters..
Here we use a modified version the reduced triplet-STDP \cite{Pfister06} but the general framework does not require this particular rule to work. 
We chose this rule since it fits experimental data well and it also offers some advantages from a computational point of view, i.e. for low firing rates it reproduces the classical STDP and for higher firing rates it is more Hebbian-like.
In the triplet-STDP, each synapse keeps track of three other values besides the synaptic weight; the synaptic traces $x_{pre}$, $x_{post_1}$ and $x_{post_2}$. 
All traces are exponentially decaying
\begin{equation}
\tau^{EE}_{x_{pre}} \dfrac{dx^{EE}_{pre}}{dt} = -x^{EE}_{pre}
\end{equation}
with the corresponding time constants $\tau^{EE}_{x_{pre}}$.
The same equation is used for ${x_{post_1}}$ and ${x_{post_2}}$ with the time constants $\tau^{EE}_{x_{post_1}}$ and $\tau^{EE}_{x_{post_2}}$, respectively.

At the arrival of a spike at the synapse the state of the synapse is updated depending on the four state variables of the synapse. 
Specifically, at the arrival of a presynaptic spike the trace $x_{pre}$, which is keeping track of the presynaptic spikes, is set to 1.
Additionally the synaptic weight change $\Delta w$ is determined using the equation
\begin{equation}
\Delta w = - \eta^{EE}_{pre} x^{EE}_{post_1} w^{\mu}
\end{equation}
where $\eta^{EE}_{pre}$ is the learning rate due to the presynaptic spike and $\mu$ is a constant which determines the dependence of the update on the previous weight.
The weight change $\Delta w$ triggered by the arrival of a postsynaptic spike is given by
\begin{equation}
\Delta w = \eta^{EE}_{post} x^{EE}_{pre} x^{EE}_{post_2} (w^{EE}_{max} - w)^{\mu}
\end{equation}
where $\eta^{EE}_{post}$ is the learning rate due to the postsynaptic spike and $w^{EE}_{max}$ is the maximum weight for a synapse between excitatory neurons. 
\textit{After} calculating the weight change, the postsynaptic traces $x^{EE}_{post_1}$ and $x^{EE}_{post_2}$ are set to 1.

The weight dependence is not included in the original triplet-STDP.
However, it is not only biologically more plausible but it is also useful to prevent weights from going to their extreme values 0 and $w^{EE}_{max}$ \cite{gutig2003}.

Note that during testing learning and homoeostasis are deactivated.
However, since learning rates are relatively low, this would only have an effect on the results when a high number of testing examples is presented (several hundred or more).

\begin{table}
	\caption{\textbf{Plasticity Parameters}}
    
	\centering
	\begin{tabular}{l l l}
		\textbf{Symbol} & \textbf{Description} & \textbf{Value} \\ \hline
		$w^{EE}_{min}$ & E to E minimum weight  & 0 \\
		$w^{EE}_{max}$ & E to E maximum weight  & 0.5 \\
		$x^{EE}_{pre}$ & E to E trace of presynaptic activity  & n.a. \\
		$x^{EE}_{post_1}$ & E to E first trace of postsynaptic activity  & n.a. \\
		$x^{EE}_{post_2}$ & E to E second trace of presynaptic activity  & n.a. \\
		$\mu$ & E to E determines weight dependence & 0.2 \\
		$\eta^{EE}_{pre}$ & E to E presynaptic spike learning rate  &  0.005 \\
		$\eta^{EE}_{post}$ & E to E postsynaptic spike learning rate  &  0.025 \\
		$\tau^{EE}_{x_{pre}}$ & E to E $x_{pre}$ time constant & 20 ms \\
		$\tau^{EE}_{x_{post_1}}$ & E to E $x_{post_1}$ time constant & 40 ms \\
		$\tau^{EE}_{x_{post_2}}$ & E to E $x_{post_2}$ time constant & 40 ms \\
		$x^{IE}_{pre}$ & I to E trace of presynaptic activity  & n.a. \\
		$x^{IE}_{post}$ & I to E trace of postsynaptic activity  & n.a. \\
		$\eta^{EE}$ & E to I learning rate  &  0.05 \\
		$\tau^{IE}_{x_{pre}}$ & I to E $x^{IE}_{pre}$ time constant & 20 ms\\
		$\tau^{IE}_{x_{post}}$ & I to E $x^{IE}_{post}$ time constant & 20 ms\\
		$\alpha$ & target firing rate & 3 Hz \\
	\end{tabular}\label{tab:plast_params}
\end{table}

\subsection{Homoeostasis}
In order to assure that each neuron is used equally and that none of the neurons is not active enough or too active we employ two homoeostatic mechanisms. 

The first mechanism is to use the learning rule presented in \cite{Vogels11}, employing plastic connections from inhibitory to excitatory neurons.
This inhibitory plasticity ensures that all excitatory neurons fire approximately as much as specified with an adjustable firing rate while maintaining an asynchronous irregular firing pattern \cite{Brunel00}.
The learning rule itself is a Hebbian STDP rule with an offset that controls the firing rate.
The synaptic state consists of three variables, namely the weight $w$, the presynaptic trace $x_{pre}$ and the postsynaptic trace $x_{post}$.
While the traces decay exponentially (as for the excitatory learning), the weight change $\Delta w$ due to a presynaptic spike is described by
\begin{equation}
\Delta w = - \eta^{IE}_{pre} (x^{IE}_{post} - 2 \alpha \tau^{IE}_{x_{post}})
\end{equation}
where $\alpha$ is the parameters controlling the postsynaptic firing rate and $\eta_{pre}$ the learning rate due to a presynaptic spike.
The weight change $\Delta w$ due to a postsynaptic spike is
\begin{equation}
\Delta w = \eta^{IE}_{post} x^{IE}_{pre}
\end{equation}
where $\eta^{IE}_{post}$ the learning rate due to a postsynaptic spike.

The second mechanism is that all weight matrices between excitatory neurons are normalized in regular time intervals using a two step procedure. 
In the first normalization step the sum of each row of each connection matrix is normalized such that it is equal to a constant.
This corresponds to a normalization of all weights along the axon of one neuron that target a certain population.
After that the sum of each column of each matrix is normalized such that it is equal to a constant.
Note that this second step ignores the normalization of the first step and will lead to non-normalized rows.
However, the sum of the rows usually stays very close to the desired constant.

As for the excitatory learning, homoeostatic mechanisms are disabled during testing.

\subsection{Input Encoding}
In this work we use rate-based \cite{london2010} population coding \cite{averbeck2006} for en-/decoding of the value represented by a neuron population.
The value represented by a neuron population is encoded by it's activity population vector \cite{Georgopoulos86} (i.e., the activity weighted sum of the preferred directions of neurons in the population).
While the previously described populations are populations of neurons, the input populations are inhomogenous Poisson spiking neurons with feedforward connections to their corresponding network populations (i.e, the neuron dynamics are not simulated, only their axonal outputs).
Each input stimulus is a Poisson spike-train with rates according to a Gaussian activation function (using wrap around).
The mean of the Gaussian represents the encoded value, as it is commonly used in population codes \cite{wu2002, pouget1997, Deneve01, Davison06, Jug12}.
The represented value of a population $\bar{a}$ is calculated using:
\begin{equation}
\bar{a} = \text{arg} \sum_{j=1}^{N^E}{\exp (i \cdot a_j)}
\end{equation}
where $a_1, ..., a_{N^E}$ are the activities of the neurons 1 to $N^E$ in the population.

\subsection{Three-way Relation Architecture}
Ultimately, we want to be able to use the previously described neuron populations as 'modules' each representing one variable of a relation. 
This relation should then be learning using the afore mentioned learning and homoeostasis mechanisms.
As an example instance of such a relational network we present a three-way relation network, relating three variables to each other.
Specifically we use one input population per variable ($X$, $Y$ and $Z$), one 'network population' per variable ($A$, $B$ and $C$) and a central network population $H$ which represents the actual relation, see figure \ref{fig:architectures}.
The neuron populations $A$, $B$ and $C$, which represent the three variables of the relation are connected to the fourth population $H$.
Note that $H$ is not connected to any one of the input populations and is therefore 'hidden' from the actual input, it only receives preprocessed information from $A$, $B$ and $C$.

As described in section \ref{sec:pop_conns}, all connections between populations are bidirectional.
Therefore (utilizing the coding scheme described in the previous subsection) we can now ensure that the inputs $X$, $Y$, $Z$ fulfil the desired input relation (e.g. $(X + Y - Z) \mod N^E = 0$) for each given example.
The connections will learn the relation and will be able to infer missing variables after learning.

The difficulty in learning such omni-directional networks is the strong feedback, since within the network there is no difference in connection strength between the connections from $H$ and to $H$.
While this sounds intuitive at first, it leads to the property that (also during learning) $H$ can have a strong impact on the activity in $A$, $B$ and $C$, thereby influencing the learning process itself.
Note that since during learning and inference the network itself is not changed, the connections from $H$ back to $A$, $B$, $C$ must be strong enough to elicit a response in the absence of their inputs.

\section{Acknowledgments}
We would like to thank you, the reader,
for reading and providing feedback on this draft!

\def\V{\rm vol.~}
\def\N{no.~}
\def\pp{pp.~}
\def\Pot{\it Proc. }
\def\IJCNN{\it International Joint Conference on Neural Networks\rm }
\def\ACC{\it American Control Conference\rm }
\def\SMC{\it IEEE Trans. Systems\rm , \it Man\rm , and \it Cybernetics\rm }

\def\handb{ \it Handbook of Intelligent Control: Neural\rm , \it
    Fuzzy\rm , \it and Adaptive Approaches \rm }

\bibliographystyle{amsplain}  
\bibliography{bib}

\end{document}